\documentclass[sigconf]{acmart}
\AtBeginDocument{%
  }

\usepackage{multirow} 
\usepackage{subcaption}
\usepackage{tcolorbox}
\usepackage{balance}
\usepackage[dvipsnames]{xcolor}
\definecolor{myblue}{RGB}{31,119,180}
\definecolor{myorange}{RGB}{255,127,14}
\definecolor{mygreen}{RGB}{44,160,44}

\copyrightyear{2026}
\acmYear{2026}
\setcopyright{cc}
\setcctype{by}
\acmConference[MM '26]{Proceedings of the 34th ACM International Conference on Multimedia}{November 10--14, 2026}{Rio de Janeiro, Brazil}
\acmBooktitle{Proceedings of the 34th ACM International Conference on Multimedia (MM '26), November 10--14, 2026, Rio de Janeiro, Brazil}
\acmDOI{10.1145/3767308.3835411}
\acmISBN{979-8-4007-2213-4/2026/11}

\begin{document}

\title{Entity-Faithful Repair of Synthetic Supervision for Zero-Shot Image Captioning}

\author{Zhiyue Liu}
\correspondingauthor
\orcid{0000-0001-6432-7836}
\affiliation{%
  \department{School of Computer, Electronics and Information}
  \department{Guangxi Key Laboratory of Multimedia Communications and Network Technology}
  \institution{Guangxi University}
  \city{Nanning}
  \state{Guangxi}
  \country{China}
}
\email{liuzhy@gxu.edu.cn}

\author{Wenkai Zhou}
\orcid{0009-0006-9319-3439}
\affiliation{%
  \department{School of Computer, Electronics and Information}
  \institution{Guangxi University}
  \city{Nanning}
  \state{Guangxi}
  \country{China}
}
\email{2413394045@st.gxu.edu.cn}
\author{Jian Qin}
\orcid{0000-0001-5411-1513}
\affiliation{%
  \department{School of Computer, Electronics and Information}
  \institution{Guangxi University}
  \city{Nanning}
  \state{Guangxi}
  \country{China}
}
\email{2413394031@st.gxu.edu.cn}

\author{Qipeng Jiang}
\orcid{0009-0001-4846-8193}
\affiliation{%
  \department{School of Computer, Electronics and Information}
  \institution{Guangxi University}
  \city{Nanning}
  \state{Guangxi}
  \country{China}
}
\email{2413394013@st.gxu.edu.cn}

\renewcommand{\shortauthors}{Zhiyue Liu, Wenkai Zhou, Jian Qin, and Qipeng Jiang}

\begin{abstract}
  Zero-shot image captioning aims to generate image descriptions without annotated image-text pairs. Recent approaches exploit text-to-image models to synthesize training data from text-only corpora, but most focus on improving overall data quality. In contrast, we observe that synthetic image-text misalignment is often structured and fine-grained: pairs may remain globally plausible while containing missing entities or misgrounded attributes, thereby degrading supervision fidelity. As a result, methods based on global similarity for image rematching or regeneration may improve apparent plausibility, but cannot systematically repair entity-level misalignment. To address this issue, we propose ReCap, a plug-and-play framework that shifts synthetic data refinement from implicit global matching to explicit fine-grained realignment. Specifically, ReCap enforces entity-level correspondence by using detected image-supported entities to guide caption rewriting, yielding more faithful synthetic supervision. In addition, we introduce an adaptive dynamic weighted learning strategy to downweight unreliable synthetic pairs during training. As a general framework, ReCap can be integrated into existing synthetic-data pipelines. Extensive experiments show that ReCap consistently improves image-text consistency and achieves state-of-the-art performance on both in-domain and cross-domain zero-shot image captioning benchmarks.
\end{abstract}




\maketitle

\section{Introduction}
Image captioning aims to generate coherent natural language descriptions for visual content. Although conventional approaches have achieved remarkable progress, they typically rely on large-scale manually annotated image-text pairs~\cite{kuo2023haav,saeidimesineh2023parallel}, whose collection is labor-intensive and expensive. Zero-shot captioning has therefore emerged as a promising direction, aiming to establish cross-modal correspondences without task-specific human annotations.

Early zero-shot paradigms~\cite{fei2023transferable,gu2023icantbelieve} are predominantly trained on text-only corpora and incorporate visual representations only at inference time. This discrepancy between training and inference fundamentally limits cross-modal grounding. More recently, advances in generative models have enabled a new paradigm: synthesizing paired training data from text-only corpora using text-to-image models such as Stable Diffusion~\cite{rombach2022high}. These approaches~\cite{liu2024improving,luo2024unleashing} alleviate the modality mismatch of text-only training by introducing synthetic image-text pairs for supervision. However, synthetic images do not always faithfully realize the semantics of their source textual prompts~\cite{kamath2023text,jin2024diagnosing,ray2023cola}. As a result, the resulting pairs often contain semantic discrepancies, such as omitted entities or misgrounded attributes, which are particularly harmful for image captioning that requires precise fine-grained alignment.

\begin{figure*}[t]
  \centering
  \includegraphics[width=0.856\textwidth]{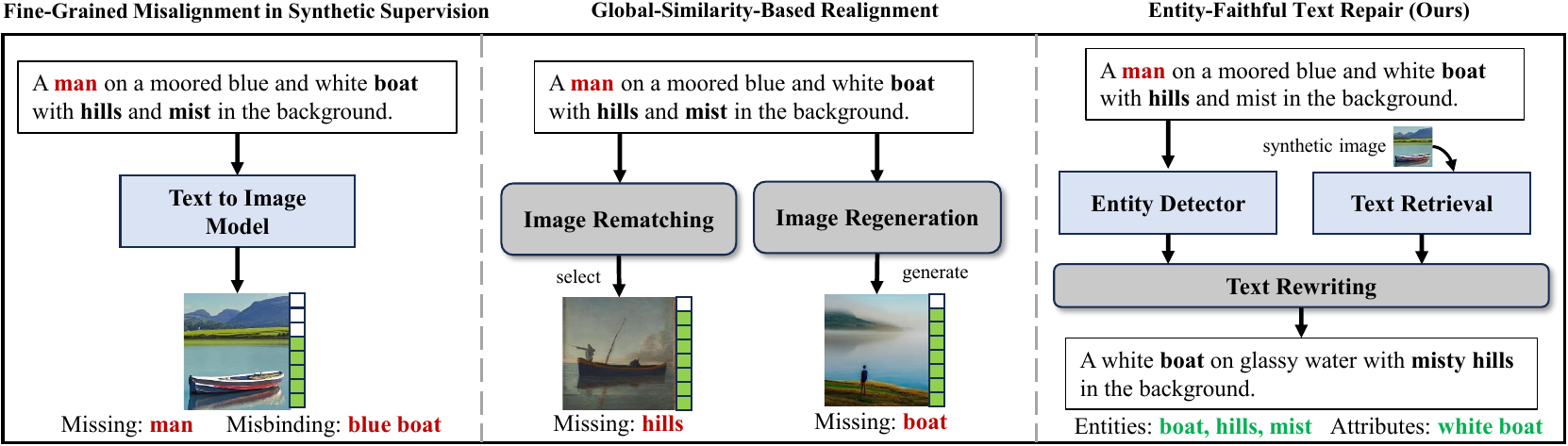}
  \caption{Illustration of structured fine-grained misalignment in synthetic image-text pairs. A pair may remain globally plausible while still containing omitted entities or misgrounded attributes. This highlights that global-similarity-based image rematching or regeneration is insufficient to correct entity-level supervision errors. In contrast, our method repairs the text supervision to be faithful to the given synthetic image.}
  \label{fig:wide_figure_11}
\end{figure*}

In the zero-shot setting, rectifying such synthetic semantic discrepancies remains challenging~~\cite{liu2025bridging}. Existing studies mainly rely on global-level repair strategies, such as image rematching~\cite{kim2025sync} or regeneration~\cite{liu2025stacap}, which search for better-aligned image--text pairs according to overall similarity. While these methods can improve apparent plausibility, they provide only coarse-grained alignment and cannot guarantee explicit correspondence at the entity level. Consequently, even highly ranked matches may still miss key entities or preserve incorrect fine-grained semantics, leaving the underlying supervision error unresolved, as illustrated in Fig.~\ref{fig:wide_figure_11}. We therefore argue that the central challenge of synthetic training data is not simply low global image--text similarity, but structured fine-grained misalignment: pairs can appear reasonable at the scene level while remaining locally misleading for supervision.

To address this issue, we propose ReCap, a novel framework for zero-shot image captioning that combines entity-guided text rewriting with adaptive dynamic-weighted training. Instead of searching for alternative images through implicit global matching, ReCap repairs the supervision text to better fit the given synthetic image. It is motivated by a simple observation: compared with synthesized images, textual descriptions are more controllable and can be revised more directly to restore faithful image-text alignment. Specifically, ReCap first identifies which prompt entities are actually supported by the synthetic image, since these entities constitute the core semantics that should be preserved in the training pair. To recover fine-grained semantics beyond simple entity presence, we further retrieve relevant textual cues from a corpus and use a large language model to rewrite the caption into an image-grounded description that better reflects visible entities, attributes, and relations. Nevertheless, even after text repair, synthetic image-text pairs may still contain residual noise. We therefore introduce an adaptive dynamic weighted learning mechanism that downweights unreliable pairs and emphasizes semantically consistent ones during training. This reliability-aware learning strategy encourages the captioning model to focus more on well-aligned synthetic pairs, improving its fine-grained cross-modal understanding.

In summary, our contributions are as follows:

\begin{itemize}
\item \textbf{Entity-Faithful Supervision Repair.} We identify that synthetic image-text misalignment in zero-shot captioning is often structured and fine-grained rather than merely globally noisy. Based on this observation, we propose an entity-guided text rewriting strategy that explicitly aligns textual supervision with image-supported entities, producing more faithful synthetic pairs.
\item \textbf{Noise-Robust Training with Adaptive Dynamic Weighting.} We introduce an adaptive dynamic-weighted learning mechanism to mitigate residual noise in synthetic supervision. By downweighting unreliable pairs during training, our method improves the robustness and fine-grained accuracy of caption generation.
\item \textbf{Comprehensive Empirical Validation.} Extensive experiments on multiple benchmarks, together with plug-and-play evaluations on existing frameworks, show that ReCap consistently improves image-text consistency and achieves state-of-the-art performance in both in-domain and cross-domain zero-shot image captioning.
\end{itemize}

\begin{figure*}[!t]
  \centering
  \includegraphics[width=0.80\textwidth]{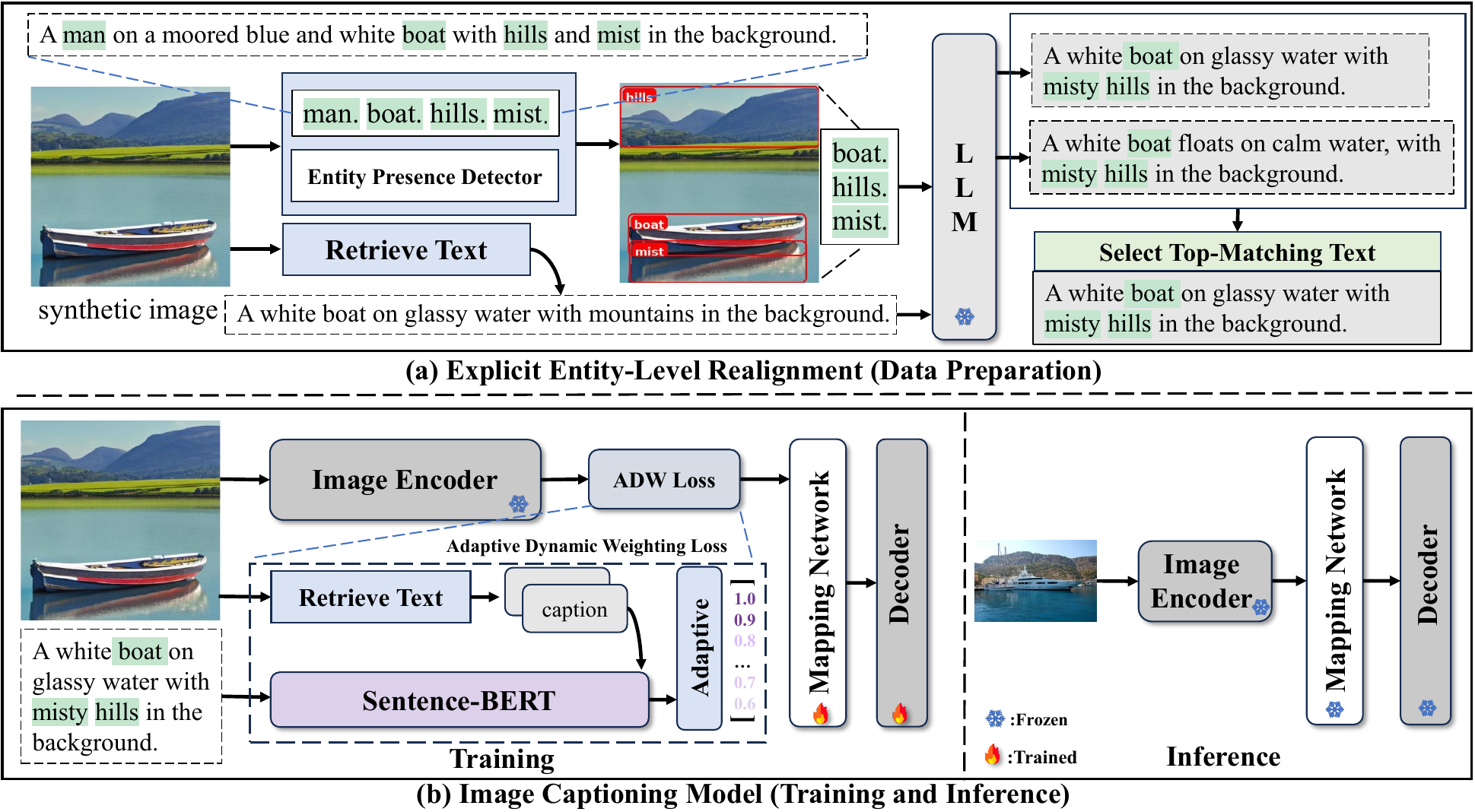}
  \caption{Overview of our proposed method. The upper part shows entity-faithful supervision repair, where image-supported entities and retrieved textual cues are used to rewrite synthetic captions into more faithful image-grounded supervision. The lower part shows reliability-aware caption training and inference, where adaptive dynamic-weighted learning (ADWL) downweights unreliable repaired pairs to improve image-text correspondence learning.}
  \label{fig:wide_figure_1}
\end{figure*}

\section{Related Work}
In contrast to supervised image captioning methods~\cite{ramos2023smallcap,kim2025vipcap}, which typically rely on manually curated image-text datasets~\cite{kuo2023haav,saeidimesineh2023parallel}, zero-shot image captioning obviates the need for task-specific human annotations. Existing zero-shot image captioning methods can be broadly classified into two paradigms: text-only approaches and synthetic-image-based approaches.

\noindent\textbf{Text-Only Methods.} 
Text-only methods train captioning models using textual corpora without paired image-text supervision, bypassing the need for paired annotations. They typically exploit the shared embedding space of pre-trained vision-language models such as CLIP~\cite{radford2021learning} to facilitate cross-modal transfer. However, these methods are fundamentally limited by the modality gap~\cite{liang2022mind}, i.e., the discrepancy between image and text representations in the shared feature space. To alleviate this issue, prior works have explored several strategies. For example, CapDec~\cite{nukrai2022text} perturbs text embeddings with Gaussian noise to better approximate image features and DeCap~\cite{li2023decap} maps image embeddings toward the corresponding textual space during inference. Entities or scene graphs can also guide caption generation~\cite{fei2023transferable,11045221}. Despite these efforts, text-only approaches still suffer from an inherent training-inference mismatch: they are optimized without image-text paired supervision during training, yet are expected to generalize to visual inputs at test time.

\noindent\textbf{Synthetic Image-Based Methods.}
To alleviate the above mismatch, a line of research synthesizes pseudo paired data from text-only corpora using text-to-image generation models~\cite{ma2024image}. By exposing captioning models to image-text pairs during training, this paradigm reduces the modality mismatch of text-only approaches and has shown clear advantages in zero-shot captioning~\cite{liu2024improving,luo2024unleashing}. However, the resulting synthetic pairs are often not perfectly aligned. Generated images may preserve the overall scene semantics while missing key entities or misgrounding fine-grained attributes. Recent studies attempt to improve synthetic data through image rematching~\cite{kim2025sync} or regeneration~\cite{liu2025stacap}, typically based on implicit global similarity. While effective to some extent, these strategies remain coarse-grained, because global similarity is not a sufficient proxy for supervision fidelity. Globally plausible candidates may still omit important entities or retain incorrect attribute grounding. Moreover, rematching-based methods are constrained by the diversity of candidate images, while regeneration-based methods may alter plausibility without guaranteeing entity-faithful correspondence.

Our work builds on synthetic image-based training, but differs from prior methods in that we target structured fine-grained misalignment in synthetic supervision. Instead of searching for globally similar alternative images, we explicitly repair the text supervision by aligning it with image-supported entities, and further reduce residual noise through adaptive dynamic-weighted learning.

\section{Methodology}

We propose ReCap, a framework for repairing synthetic supervision in zero-shot image captioning. As illustrated in Fig.~\ref{fig:wide_figure_1}, ReCap addresses synthetic image-text misalignment in two stages. First, it performs entity-faithful supervision repair by rewriting the text side of a synthetic pair according to the entities that are visually supported by the generated image (Sec.~\ref{sec:3.1}). Second, since residual noise may still remain after rewriting, it adopts an adaptive dynamic-weighted learning strategy to reduce the influence of unreliable synthetic pairs during training (Sec.~\ref{sec:3.2}). The resulting repaired supervision is then used to train a lightweight image captioning model for zero-shot caption generation (Sec.~\ref{sec:3.3}).

\subsection{Explicit Entity-Level Realignment}
\label{sec:3.1}
Given a text corpus $\mathcal{D}_t = \{t_1, \ldots, t_n\}$ containing $n$ entries, we leverage a pre-trained text-to-image generative model $G$ to synthesize an image for each text:
\begin{equation}
I_i = G(t_i), \quad i = 1, \ldots, n.
\end{equation}

This would result in a set of initial pseudo paired data $\mathcal{D}_p = \{(I_1, t_1), \ldots, (I_n, t_n)\}$. However, training an image captioning model directly on $\mathcal{D}_p$ yields suboptimal performance. This is primarily due to the inherent structured fine-grained misalignment within the synthesized pairs, which introduces erroneous cross-modal supervision and weakens the fine-grained alignment between visual and textual representations. To rectify this, we propose an entity-guided text rewriting strategy that explicitly repairs the text supervision to better match the given synthetic image.

We first determine which entities mentioned in the original text are actually supported by the synthesized image. Specifically, a text parser $\mathcal{P}$ is used to extract the entity set from an original caption $t$ as $E  = \mathcal{P}(t)=\{e_1,\ldots,e_m\}.$ All extracted entities are then matched against the corresponding image $I$ using an entity presence detector $\mathcal{O}$, yielding the set of entities actually present in the image:
$
E' = \mathcal{O}(E, I)=\{e'_1,\ldots,e'_{m'}\}, E' \subseteq E.
$
This step removes unsupported entities from the textual supervision and directly addresses entity omission in synthetic image-text pairs.

While entity filtering mitigates omission errors, it cannot fully prevent incorrect attribute grounding, because retained entities may still be assigned properties unsupported by the image. To introduce richer and more reliable image-grounded semantic cues, we retrieve supplementary textual information that is semantically relevant to the content of the synthesized image. Specifically, we encode the image into a CLIP visual embedding:  

\begin{equation}
\mathbf{v} = f_{\text{img}}(I) \in \mathbb{R}^d.
\label{eq:2}
\end{equation}

Because the original source text may itself contain unsupported or misleading content, using it directly for scene interpretation can propagate errors. We therefore use image embeddings to retrieve the most semantically relevant text from the entire corpus, which not only provides complementary contextual information but also facilitates the reconstruction of accurate scene understanding:
\begin{equation}
t' = \arg\max_{t_i \in \mathcal{D}_t} 
\cos\!\left( \mathbf{v},\, f_{\text{text}}(t_i) \right).
\label{eq:3}
\end{equation}
The retrieved text serves as complementary textual evidence for reconstructing more faithful scene semantics. We then concatenate the image-supported entity set $E'$, the retrieved text $t'$, and a rewriting prompt, and feed them into a large language model $\mathcal{L}$ to generate candidate repaired captions as follows:
\begin{equation}
\tilde{T} = \{\tilde{t}_1,\ldots,\tilde{t}_k\}
    = \mathcal{L}\bigl(\text{concat}(\text{prompt},\, E',\, t',\, k )\bigr),
\end{equation}
where $k$ denotes the number of candidate rewrites. Since the outputs of $\mathcal{L}$ may vary in quality, we adopt a generate-then-align strategy to select the final repaired caption. Specifically, we augment the candidate set with the original text: $\tilde{T'} = \{t,\; \tilde{t}_1,\; \ldots,\; \tilde{t}_k\}$, and rank all candidates according to their CLIP similarity to the synthesized image as follows:
\begin{equation}
\tilde{t} = \arg\max_{\tilde{t}_i \in \tilde{T'} }
\cos\!\left( \mathbf{v},\, f_{\text{text}}(\tilde{t}_i) \right).
\label{eq:6}
\end{equation}
Including the original text in the candidate pool allows the method to preserve the original supervision when rewriting does not provide a better alignment.

Finally, we obtain a repaired synthetic dataset:
\begin{equation}
\mathcal{D}_{\text{rp}}
= \{(I_1, \tilde{t}_1), \ldots, (I_n, \tilde{t}_n)\}.
\end{equation}
Compared with the original pseudo pairs, $\mathcal{D}_{\text{rp}}$ provides more faithful text supervision for caption training by reducing entity omission and improving fine-grained semantic grounding.
\subsection{Adaptive Dynamic Weighting Loss}
\label{sec:3.2}
Although the rewriting step improves entity-level faithfulness, synthetic supervision may still contain residual noise. For example, the repaired caption may remain semantically incomplete or overly generic, and the retrieved textual cues may not perfectly capture all scene details. To make training robust to such residual noise, we introduce an adaptive dynamic-weighted learning strategy that assigns lower weights to less reliable synthetic pairs.

Given a repaired pair $(I_i,\tilde{t}_i)\in \mathcal{D}_{\text{rp}}$, we estimate its semantic reliability by comparing the rewritten caption with a set of retrieved textual references. Specifically, for image $I_i$, we retrieve the top-$l$ most relevant captions from the corpus and form a semantic reference set: $T_r^{(i)} = \{t'^{(i)}_1,\ldots,t'^{(i)}_l\}$. Since both $\tilde{t}_i$ and $T_r^{(i)}$ are textual, measuring consistency in text space avoids introducing additional uncertainty from visual encoding and provides a direct estimate of whether the rewritten caption preserves the main image-grounded semantics. We use Sentence-BERT~\cite{reimers-2019-sentence-bert} as a sentence-level semantic encoder, denoted by $E_S(\cdot)$, and compute the average semantic consistency score as follows:
\begin{equation}
s_i = \frac{1}{l}\sum_{j=1}^{l}
\cos\!\left(E_S(\tilde{t}_i),\, E_S(t'^{(i)}_j)\right).
\end{equation}
A larger $s_i$ indicates that the rewritten caption is more consistent with the retrieved semantic references and is therefore more likely to provide reliable supervision. We then convert $s_i$ into a sample weight $w_i$ and use it to reweight the captioning loss. In practice, the weight is calculated by:
\begin{equation}
w_i = \phi(s_i) = \min\{1, \max\{0, c*s_i\}\},
\end{equation}
where $c$ is a scaling parameter. Capping the maximum weight can prevent a few high-scoring samples from dominating optimization. We denote the sequence-level negative log-likelihood loss as $\mathcal{L}^{(i)}_{\text{cap}}$ for the $i$-th training pair as follows:
\begin{equation}
\mathcal{L}^{(i)}_{\text{cap}}
= - \sum_{t=1}^{T_i}\log p(y_{i,t}\mid y_{i,<t}, I_i),
\end{equation}
where $T_i$ is the caption length and $y_{i,t}$ is the target token at step $t$. The final adaptive dynamic-weighted loss is defined as:
\begin{equation}
\mathcal{L}_{\text{ADW}}
=
\frac{1}{B}\sum_{i=1}^{B} w_i \, \mathcal{L}^{(i)}_{\text{cap}},
\label{eq:adwl}
\end{equation}
where $B$ denotes the batch size. In this way, pairs with higher semantic consistency receive stronger supervisory weights, while less reliable synthetic pairs are downweighted. This improves robustness to residual noise and stabilizes caption learning from repaired synthetic supervision.

\subsection{Image Captioning Architecture}
\label{sec:3.3}

After repairing synthetic supervision, we use the resulting dataset $\mathcal{D}_{\text{rp}}$ to train a lightweight image captioning model. Given a repaired pair $(I,\tilde{t})$, we first extract the visual embedding $\mathbf{v}$ using the visual encoder defined in Eq.~\ref{eq:2}. The visual feature is then fed into a lightweight mapping network, which projects it into the input embedding space of the caption decoder. Finally, the decoder generates the target caption $\tilde{t}$ in an autoregressive manner.

The model is trained with the adaptive dynamic-weighted loss in Eq.~\ref{eq:adwl}, so that more reliable repaired pairs contribute more strongly to optimization. At inference time, only the image captioning model is used, without any retrieval or rewriting modules.

\section{Experimentation}
In this section, we evaluate ReCap’s effectiveness in addressing fine-grained misalignment in synthetic supervision, conducting experiments on standard image captioning datasets across in-domain, cross-domain, and other zero-shot settings to assess both its captioning performance and its generalization ability.
\begin{table*}[t]
\caption{In-domain captioning results on MSCOCO and Flickr30k datasets. The best results are highlighted in bold. ``-'' indicates that the metric is not reported.}
\centering
\resizebox{0.63\textwidth}{!}{
\begin{tabular}{l|cccc|cccc}
\toprule
\multirow{2.5}{*}{\textbf{Method}}
& \multicolumn{4}{c|}{\textbf{MSCOCO}} 
& \multicolumn{4}{c}{\textbf{Flickr30k}} \\
\cmidrule(lr){2-5} \cmidrule(lr){6-9}
& \textbf{B@4} & \textbf{M} & \textbf{C} & \textbf{S} 
& \textbf{B@4} & \textbf{M} & \textbf{C} & \textbf{S} \\
\midrule
\multicolumn{9}{l}{\textit{without inference-time retrieval}} \\
CapDec~\cite{nukrai2022text}    \textcolor{gray}{EMNLP'22} & 26.4 & 25.1 & 91.8  & {\text{-}}
& 17.7 & 20.0 & 39.1 & - \\
DeCap~\cite{li2023decap}      \textcolor{gray}{ICLR'23}& 24.7 & 25.0 & 91.2 &  18.7
& 21.2 & \textbf{21.8} &  56.7 &  15.2 \\
C3~\cite{zhang2024connect}  \textcolor{gray}{ICLR'24}& 27.7 & 25.0 & 93.3 & 18.3  
& - & - & - & - \\
Diffusion Bridge~\cite{11094818}   \textcolor{gray}{CVPR'25}  &  28.7 &  25.1 &  96.4 &  18.7 
&  21.4 & 21.1 & 53.0 & 14.5 \\
ReCap   \textcolor{gray}{Ours}& \textbf{29.3} & \textbf{25.2} & \textbf{98.5} & \textbf{18.8} 
& \textbf{23.6} & 21.5 & \textbf{56.8} & \textbf{15.4} \\
\midrule
\multicolumn{9}{l}{\textit{with retrieval}} \\
VIECap~\cite{fei2023transferable}   \textcolor{gray}{ICCV'23}  & 27.2 & 24.8 & 92.9 & 18.2 & 21.4 & 20.1 & 47.9 & 13.6 \\
MeaCap~\cite{zeng2024meacap}      \textcolor{gray}{CVPR'24} &27.2 &25.3& 95.4& 19.0& 22.3& 22.3& 59.4& 15.6 \\
SynTIC~\cite{liu2024improving}     \textcolor{gray}{AAAI'24}& 29.9 & 25.8 & 101.1 & 19.3 & 22.3 & 22.4 & 56.6 & 16.6\\
PCM-Net~\cite{luo2024unleashing}    \textcolor{gray}{ECCV'24} & 31.5 & 25.9 & 103.8 & 19.7 &  26.9 &  23.0 & 61.3 & 16.8 \\
IFCap~\cite{lee2024ifcap}          \textcolor{gray}{EMNLP'24} & 30.8 &  26.7 &  108.0 &  20.3 & 23.5 &  23.0 &  64.4 &  17.0\\
SaCap~\cite{liu2025stacap}  \textcolor{gray}{KBS'25}&31.0 &25.9  &104.2 &19.7 &\textbf{27.1} &22.4  &64.3 &15.9\\
SynC~\cite{kim2025sync}     \textcolor{gray}{ACMMM'25}    & \textbf{31.9} & 25.7 & 105.9  & 19.6 & - & - & - & - \\
SRACap~\cite{11045221}             \textcolor{gray}{TPAMI'25}  & 27.8 & 25.3 & 92.6  & 18.3 &20.5&20.7&47.4&14.7 \\
MERCap~\cite{zeng2025zero}        \textcolor{gray}{AAAI'25}& 27.3&25.5&96.0&19.5&23.2&22.3&57.2&15.9\\
ReCap+retrieval \textcolor{gray}{Ours}   & 31.3 & \textbf{26.8}&\textbf{110.5}&\textbf{20.4} &
24.1& \textbf{23.1} & \textbf{65.2}  &\textbf{17.4} \\ 
\bottomrule
\end{tabular}
}
\label{tab:indomain_combined}
\end{table*}

\subsection{Experimental Settings}

\begin{table}[t]
\centering
\caption{Extensibility on different captioning frameworks.}
\resizebox{0.48\textwidth}{!}{
\begin{tabular}{l|cccc|cccc}
\toprule
\multirow{2.5}{*}{\textbf{Method}} 
& \multicolumn{4}{c|}{\textbf{CapDec}}
& \multicolumn{4}{c}{\textbf{VIECap}} \\
\cmidrule(lr){2-5}
\cmidrule(lr){6-9}

& \textbf{B@4} & \textbf{M} & \textbf{C} & \textbf{S}
& \textbf{B@4} & \textbf{M} & \textbf{C} & \textbf{S} \\
\midrule
Baseline
& 26.4 & 25.1 & 91.8 & --
& 27.2 & 24.8 & 92.9 & 18.2 \\

{\textit{+SynC}}
& 27.7 & \textbf{25.7} & 96.4 & 19.3
& 27.7 & 24.8 & 96.4 & 18.5 \\

{\textit{+ReCap}}
& \textbf{29.1} & 25.4 & \textbf{97.7} & \textbf{19.5}
& \textbf{29.0} & \textbf{25.2} & \textbf{98.0} & \textbf{18.7} \\

\midrule
\multirow{2.5}{*}{\textbf{Method}} 
& \multicolumn{4}{c|}{\textbf{PCM-Net}}
& \multicolumn{4}{c}{\textbf{IFCap}} \\
\cmidrule(lr){2-5}
\cmidrule(lr){6-9}
& \textbf{B@4} & \textbf{M} & \textbf{C} & \textbf{S}
& \textbf{B@4} & \textbf{M} & \textbf{C} & \textbf{S} \\
\midrule
Baseline
& 31.5 & \textbf{25.9} & 103.8 & 19.7
& 30.8 & 26.7 & 108.0 & 20.3 \\

{\textit{+SynC}}
& 31.9 & 25.7 & 105.9 & 19.6
& 31.1 & 26.4 & 108.6 & 20.3 \\

{\textit{+ReCap}}
& \textbf{32.1} & \textbf{25.9} & \textbf{106.8} & \textbf{20.1}
& \textbf{31.2} & \textbf{26.9} & \textbf{110.2} & \textbf{20.6} \\
\bottomrule
\end{tabular}
}
\label{tab:Extensibility}
\end{table}
\noindent\textbf{Datasets.} The experiments are conducted on three widely used image captioning benchmarks, including MSCOCO~\cite{chen2015coco}, Flickr30k~\cite{young2014visual}, and NoCaps~\cite{agrawal2019nocaps}. For MSCOCO and Flickr30k, we follow the Karpathy split~\cite{karpathy2015deep}. For NoCaps, we follow its official evaluation split.

\noindent\textbf{Metrics. }
We report four standard image captioning metrics, including CIDEr (\textbf{C})~\cite{vedantam2015cider}, SPICE (\textbf{S})~\cite{anderson2016spice}, BLEU-4 (\textbf{B@4})~\cite{papineni2002bleu}, and METEOR (\textbf{M})~\cite{banerjee2005meteor}. Among them, CIDEr is the primary metric, as it is widely used for caption generation and better reflects agreement with multiple human references~\cite{rennie2017self}.

\noindent\textbf{Implementation Details.} For image synthesis, we use Stable Diffusion v1.5 to generate images at a resolution of $512 \times 512$ with 20 sampling steps, while textual descriptions are rewritten using the large language model Llama 3-8B~\cite{grattafiori2024llama3}. Grounding DINO~\cite{liu2024groundingdinomarryingdino} is employed for entity presence detection. Notably, Grounding DINO relies on predefined entity queries and is used only to verify whether text-mentioned entities appear in the image, without adding extra visual recognition capacity, ensuring a fair comparison. NLTK~\cite{bird2004nltk} is used for text parsing. Our model backbone adopts CLIP ViT-B/32 to encode both images and texts, with the base version of GPT-2-base~\cite{radford2019language} as the language decoder. 
Visual and textual representations are aligned through a mapping network implemented as an 8-layer Transformer. During training, all trainable parameters are optimized using AdamW~\cite{loshchilov2019decoupled} with a learning rate of $2 \times 10^{-5}$ and a batch size of 40. 
We apply a linear learning rate warm-up over the first 5,000 steps.
All experiments are conducted on a single NVIDIA RTX 3090 GPU.

\noindent\textbf{Baselines.} Our method does not rely on the external memory bank or object detector during inference. Rather than relying on retrieval-based mappings for caption generation, it benefits from carefully curated training data. We compare ReCap with the methods without inference-time retrieval, including CapDec~\cite{nukrai2022text}, DeCap~\cite{li2023decap}, C3~\cite{zhang2024connect}, and Diffusion Bridge~\cite{11094818}, as well as the retrieval-based methods that leverage external memory during inference, such as VIECap~\cite{fei2023transferable}, MeaCap~\cite{zeng2024meacap}, SynTIC~\cite{liu2024improving}, PCM-Net~\cite{luo2024unleashing}, IFCap~\cite{lee2024ifcap}, SRACap~\cite{11045221}, MERCap~\cite{zeng2025zero}, SaCap~\cite{liu2025stacap}, and SynC~\cite{kim2025sync}. For SynC, we report the CLIP ViT-B/32 version to ensure a fair comparison under the same visual-language backbone.

\subsection{In-Domain Zero-Shot Captioning}
\noindent\textbf{Results without Inference-Time Retrieval. } Table~\ref{tab:indomain_combined} reports the in-domain zero-shot captioning results on MSCOCO and Flickr30k. Under the setting without inference retrieval, ReCap achieves the best performance on all metrics on MSCOCO, demonstrating that repairing synthetic supervision alone can already provide highly effective training signals for zero-shot captioning. On Flickr30k, ReCap remains highly competitive, attaining state-of-the-art scores on BLEU-4, CIDEr, and SPICE, and achieving the second-highest score on METEOR. This shows that explicitly repairing structured fine-grained misalignment in synthetic pairs leads to more faithful supervision and stronger caption generation, even without relying on external memory or retrieval at inference time.

\begin{table*}[t]
\caption{Cross-domain zero-shot captioning results, where X$\rightarrow$Y means source domain$\rightarrow$target domain.}
\centering
\resizebox{0.71\textwidth}{!}{
\begin{tabular}{l|cccc|cccc|cccc}
\toprule
\multirow{2.5}{*}{\textbf{Method}} &
\multicolumn{4}{c|}{\textbf{MSCOCO$\rightarrow$Flickr30k}} &
\multicolumn{4}{c|}{\textbf{Flickr30k$\rightarrow$MSCOCO}} &
\multicolumn{4}{c}{\textbf{MSCOCO$\rightarrow$NoCaps}} \\
\cmidrule(lr){2-5} \cmidrule(lr){6-9} \cmidrule(lr){10-13}
& \textbf{B@4} & \textbf{M} & \textbf{C} & \textbf{S} 
& \textbf{B@4} & \textbf{M} & \textbf{C} & \textbf{S} 
& \textbf{In} & \textbf{Near} & \textbf{Out} & \textbf{All} \\
\midrule
CapDec~\cite{nukrai2022text}           & 17.3 & \textbf{18.6} & 35.7 & -- & 9.2 & 16.3 & 27.3 & --&60.1& 50.2& 28.7&45.9\\
DeCap~\cite{li2023decap}            & 16.3 &  17.9 & 35.7 & 11.1 &12.1 &  18.0 & 44.4 &  10.9
& 65.2&47.8&25.8& 45.9 \\
Diffusion Bridge~\cite{11094818} &  17.4 & \textbf{18.6} &  38.5 &  11.2 & 14.7 &  18.0 &  47.1 & \textbf{12.0}
& --& --& -- & -- \\
ReCap
& \textbf{18.3} & \textbf{18.6} & \textbf{41.5} & \textbf{12.2}
& \textbf{14.8} & \textbf{18.3} & \textbf{49.9} & \textbf{12.0}	
& \textbf{68.4}& \textbf{63.1}& \textbf{43.3}& \textbf{58.9}\\

\bottomrule
\end{tabular}
}
\label{tab:cross_domain}
\end{table*}

\noindent\textbf{Results with Inference-Time Retrieval.} Since ReCap does not perform inference retrieval and directly maps images to text, directly comparing it with retrieval-based methods is not entirely fair. To evaluate performance under the retrieval setting, we combine our repaired synthetic supervision with a lightweight retrieval module inspired by IFCap, denoted as ReCap+retrieval. As shown in Table~\ref{tab:indomain_combined}, our method achieves state-of-the-art results of METEOR, CIDEr, and SPICE on MSCOCO, while ranking third on BLEU-4. In particular, it improves CIDEr by 2.5 points over IFCap, indicating that our repaired image-text pairs provide a more reliable basis for retrieval-based generation. On Flickr30k, our method also achieves the best results on METEOR, CIDEr, and SPICE, further confirming that our supervision repair strategy remains effective when combined with inference-time retrieval.

\noindent\textbf{Plug-and-Play Extensibility.} We further evaluate the extensibility of ReCap by integrating it into several existing zero-shot captioning frameworks. Specifically, we adapt text-only trained models, including CapDec, VIECap, and IFCap, so that they can be trained on synthetic image-text pairs. For CapDec and VIECap, the CLIP text encoder is replaced with the CLIP image encoder, and the noise injection step is removed. For IFCap, its image-style retrieval procedure is substituted with a standard image-to-text retrieval mechanism. In addition, we directly leverage ReCap's refined data to train PCM-Net. Table~\ref{tab:Extensibility} reports the in-domain performance on MSCOCO. Compared to SynC, ReCap consistently enhances the performance of multiple zero-shot captioning models, further demonstrating its broad applicability and confirming that the refined image-text pairs are indeed more consistent.

\subsection{Cross-Domain Zero-Shot Captioning}
Cross-domain experiments evaluate the generalization capability of captioning models by requiring them to generate captions in a target domain after being trained on a different source domain. We consider three settings: MSCOCO $\rightarrow$ Flickr30k, Flickr30k $\rightarrow$ MSCOCO, and MSCOCO $\rightarrow$ NoCaps, as shown in Table~\ref{tab:cross_domain}.

Since retrieval-based methods may implicitly access target-domain information through retrieved entities or captions, direct comparison with such methods is less controlled in the cross-domain setting. Moreover, their performance is strongly coupled with the retrieval pipeline itself. We therefore report the base model ReCap, without inference-time retrieval, as the primary comparison in order to better evaluate intrinsic cross-domain generalization.

Under the MSCOCO $\rightarrow$ Flickr30k setting, our method achieves strong performance on all four metrics and improves CIDEr by 3.0 points over Diffusion Bridge. This result suggests that repairing structured fine-grained misalignment in synthetic supervision helps the model preserve more reliable visual-textual correspondences under domain shift, leading to captions with better object, attribute, and relation grounding.

Under the Flickr30k $\rightarrow$ MSCOCO setting, ReCap obtains the best overall performance, with a CIDEr gain of 2.8 points over Diffusion Bridge. This demonstrates that the benefit of supervision repair is not limited to a specific source domain. Instead, by improving the faithfulness of synthetic training pairs, ReCap yields more robust cross-domain transfer to target distributions with broader semantic diversity.

For the MSCOCO $\rightarrow$ NoCaps benchmark, In, Near, Out, and All denote the in-domain, near-domain, out-of-domain, and overall splits, respectively, and all results are reported in terms of CIDEr. ReCap achieves the best performance on all splits without inference-time retrieval. Since NoCaps contains more diverse and out-of-domain visual concepts, these gains further demonstrate that repairing synthetic supervision at a fine-grained level improves robustness beyond the training distribution. Overall, the cross-domain results show that ReCap improves zero-shot captioning not only in the in-domain setting, but also under substantial domain shift.

Relying solely on visual-textual alignment without retrieval or object detectors, these gains show that improving synthetic data faithfulness enables robust zero-shot captioning across domains.

\begin{table}[t]
\caption{Ablation results for each component.}
\centering
\begin{tabular}{l|cccc}
\toprule
\textbf{Method} &  \textbf{B@4} & \textbf{M} & \textbf{C} & \textbf{S}  \\
\midrule
Baseline          &26.1&24.4&89.8&18.0 \\
{\textit{+ETR}}   &27.7&25.2&95.6&18.6 \\
{\textit{+ADWL}}   &27.8&25.2&96.4&18.7 \\
ReCap            & \textbf{29.3} & \textbf{25.2} & \textbf{98.5}  & \textbf{18.8} \\
\bottomrule
\end{tabular}
\label{tab:Components}
\end{table}

\begin{table}[t]
\caption{Effect of the ADWL scaling $c$, number of generated texts $k$, size of the semantic reference set $l$, and number of retrieved texts $m$.}
\centering
\begin{tabular}{cc|cc|cc|cc}
\toprule
$c$ & \textbf{CIDEr} & $l$ & \textbf{CIDEr} & $k$ &\textbf{CIDEr} & $m$ &\textbf{CIDEr}\\
\midrule
1.0 & 97.7 & 1 & 97.6 & 1 & 98.5 & 0& 89.8  \\
1.2 & 98.0 & 2 & 98.0 & 2 & 98.4 & 1& \textbf{95.6}  \\
1.4 & 98.3 & 3 & \textbf{98.5} & 3 & 98.0 &2 &93.4 \\
1.6 & \textbf{98.5} & 4 & 98.3 & 4 & \textbf{98.6} &3 & 91.8\\
1.8 & 98.1 & 5 & 98.1 & 5 & 98.2 &4 & 92.3\\
\bottomrule
\end{tabular}
\label{tab:hyper}
\end{table}

\subsection{Ablation Studies}
\noindent\textbf{Effect of Components.}
To validate the effectiveness of ReCap, we conduct ablation studies on the MSCOCO dataset. We first establish a baseline that trains the captioning model on synthetic image–text pairs without pair refinement and performs decoding solely based on image grid features. We then progressively incorporate two key components into the baseline: the entity-guided text rewriting (ETR) paradigm and the adaptive dynamic weighting loss (ADWL). As shown in Table~\ref{tab:Components}, each component consistently leads to performance improvements.

\noindent\textbf{Effect of ADWL Scaling.}
We analyze the scaling parameter $c$ in ADWL, which controls how strongly semantic reliability affects the sample weight. The best performance is achieved at $c=1.6$, and larger values tend to overemphasize low-quality pairs, leading to performance degradation, as shown in Table~\ref{tab:hyper}. This supports the role of ADWL as a moderate reliability-aware reweighting strategy.

\noindent\textbf{Effect of Semantic Reference Numbers.}
We also examine the number of retrieved semantic reference texts $l$ used in ADWL. Table~\ref{tab:hyper} shows that using only one reference is insufficient to capture the main semantics of the image-text pair, while the best performance is achieved at $l=3$. Further increasing $l$ introduces additional noisy references, which weakens the reliability estimate and slightly degrades performance. This suggests that ADWL benefits from a small but diverse set of semantic references.

\noindent\textbf{Effect of Generated Text Numbers.}
We further study the number of candidate rewrites $k$ under the MSCOCO in-domain setting. As shown in Table~\ref{tab:hyper}, increasing $k$ does not bring further gains. This is consistent with the design of ReCap: the rewriting stage is mainly constrained by the same image-supported entities and retrieved textual cues, so generating more candidates brings limited additional semantic benefit. We therefore set $k=1$ for efficiency.

\noindent\textbf{Effect of Retrieved Texts for Rewriting.}
We extend the retrieval step in Eq.~\ref{eq:3} to select the top-$m$ texts ranked by their CLIP similarity to the synthetic image and concatenate them as complementary cues for rewriting. The setting $m=0$ removes retrieved textual cues. As shown in Table~\ref{tab:hyper}, retrieving multiple texts degrades performance, likely because additional captions introduce redundant or conflicting semantic cues. We therefore use $m=1$ in experiments.

Overall, ETR repairs entity-level and fine-grained semantic misalignment, while ADWL improves robustness by suppressing residual noisy pairs.

\subsection{Study on the Repair Process}
\noindent\textbf{Analysis of Image-Text Consistency.}
To evaluate whether ReCap improves the fidelity of synthetic supervision, we measure the quality of repaired synthetic image-text pairs constructed from the MSCOCO and Flickr30k training sets. We adopt two commonly used image-text consistency metrics, CLIPScore~\cite{DBLP:journals} and MMScore~\cite{kim2025sync}, and report the results in Table~\ref{tab:Consistency}. Compared with the original synthetic pairs, the repaired pairs consistently achieve higher scores on both metrics. This indicates that ReCap does not merely rewrite captions superficially, but produces image-text pairs with stronger semantic alignment and more faithful supervision for training.

\begin{table}[t]
\caption{Text–image consistency results.}
\centering
\begin{tabular}{l|cc|cc}
\toprule
\multirow{2}{*}{\textbf{Dataset}} 
& \multicolumn{2}{c|}{\textbf{CLIPScore}} 
& \multicolumn{2}{c}{\textbf{MMScore}} \\
\cmidrule(lr){2-3} \cmidrule(lr){4-5}
& \textbf{MSCOCO} & \textbf{Flickr30k} 
& \textbf{MSCOCO} & \textbf{Flickr30k} \\
\midrule

Original  & 30.92 & 30.43 & 66.57 & 76.95 \\
Repaired & \textbf{31.67} & \textbf{32.30} & \textbf{77.52} & \textbf{82.31} \\

\bottomrule
\end{tabular}
\label{tab:Consistency}
\end{table}

\begin{figure}[!t]
  \begin{subfigure}[b]{0.23\textwidth}
    \centering
    \includegraphics[width=\linewidth]{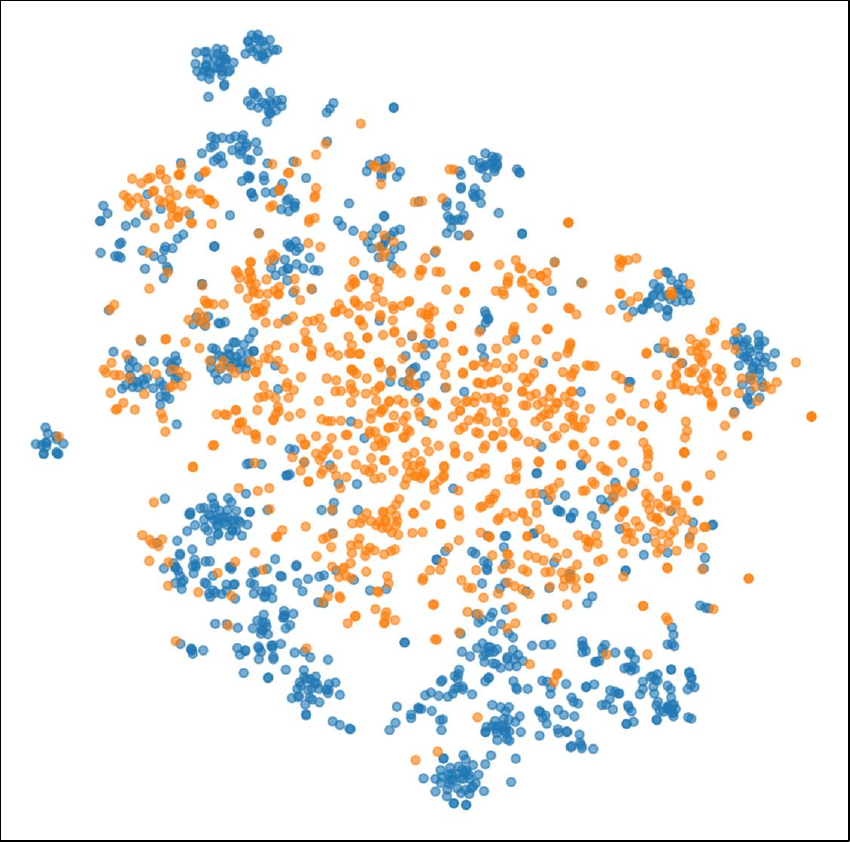}
    \label{fig:wide_figure_5_1}
    \caption{MSCOCO: \textcolor{myorange}{$\bullet$} vs. Flickr30k: \textcolor{myblue}{$\bullet$}} 
  \end{subfigure}
  \hfill
  \begin{subfigure}[b]{0.23\textwidth}
    \centering
    \includegraphics[width=\linewidth]{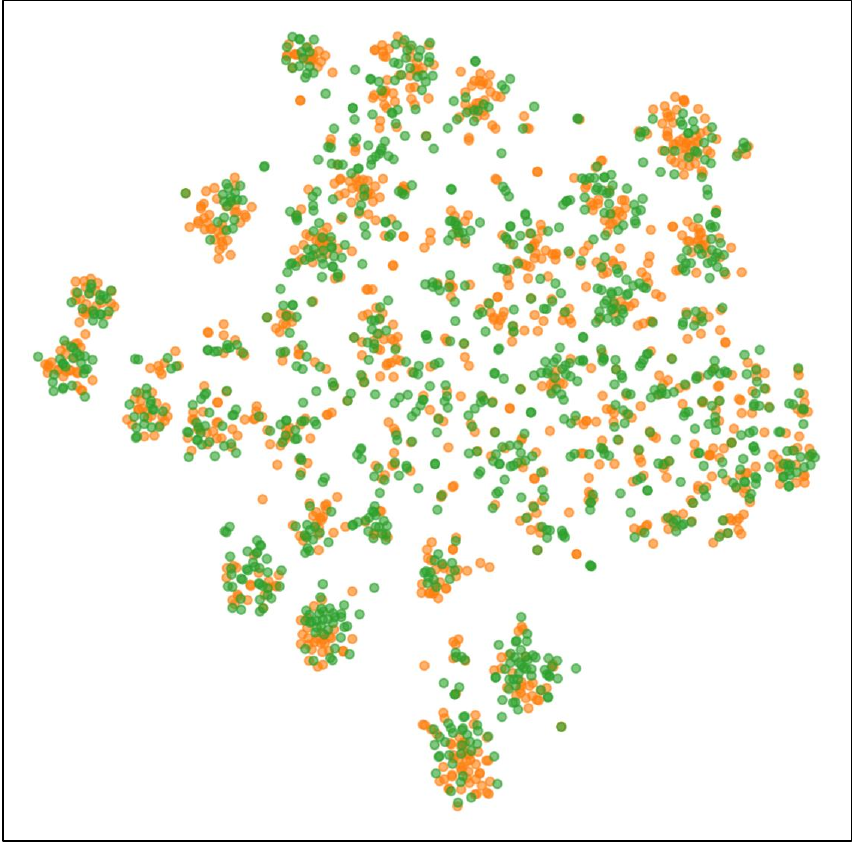}
    \label{fig:wide_figure_5_2}
    \caption{Original: \textcolor{myorange}{$\bullet$} vs. Rewritten: \textcolor{mygreen}{$\bullet$} }
  \end{subfigure}

  \caption{Text embedding distributions.}
  \label{fig:wide_figure_5}
\end{figure}

\begin{table}[t]
\centering
\caption{Robustness to false negatives in entity verification.
Entity deletion denotes the proportion of correctly verified entities
that are randomly removed.}
\label{tab:entity_ablation}
\resizebox{0.48\textwidth}{!}{
\begin{tabular}{c|cccc|cccc}
\toprule
\multirow{2.5}{*}{\textbf{Deletion} (\%)}
& \multicolumn{4}{c|}{\textbf{without ADWL}}
& \multicolumn{4}{c}{\textbf{with ADWL}} \\
\cmidrule(lr){2-5}
\cmidrule(lr){6-9}
& \textbf{B@4} & \textbf{M} & \textbf{C} & \textbf{S}
& \textbf{B@4} & \textbf{M} & \textbf{C} & \textbf{S} \\
\midrule

100
& 26.1 & 24.4 & 89.8 & 18.0
& 27.8 & \textbf{25.2} & 96.4 & 18.7 \\

50
& 27.3 & 24.4 & 92.1 & 17.9
& 28.0 & 25.1 & 96.9 & 18.7 \\

0
& \textbf{27.7} & \textbf{25.2} & \textbf{95.6} & \textbf{18.6}
& \textbf{29.3} & \textbf{25.2} & \textbf{98.5} & \textbf{18.8} \\
\bottomrule
\end{tabular}
}
\end{table}

\begin{table}[t]
\centering
\caption{Entity recall of different methods. ReCap$^{*}$ denotes our method using real images.}
\label{tab:entity_recall}
\resizebox{0.45\textwidth}{!}{
\begin{tabular}{lccccc}
\toprule
\textbf{Method}
& \textbf{Baseline} 
& \textbf{CapDec}
& \textbf{DeCap}
& \textbf{ReCap$^{*}$}
& \textbf{ReCap} \\
\midrule
\textbf{Recall}
&0.3399
& 0.3419
& 0.3345
& \textbf{0.3666}
& 0.3510 \\
\bottomrule
\end{tabular}
}
\end{table}

\begin{table}[t]
\centering
\caption{Time cost comparison involving image synthesis.}
\label{tab:time_cost}
\begin{tabular}{llccc}
\toprule
\textbf{Dataset} & \textbf{Time Cost} & \textbf{ReCap} & \textbf{SynC} & \textbf{SaCap} \\
\midrule
\multirow{2}{*}{MSCOCO}
    & Total time (h)     & 159.5 & 141.3 & 549.0 \\
    & Per-image time (s) & 1.01  & 0.90  & 3.49  \\
\bottomrule
\end{tabular}
\end{table}

\begin{figure*}[t]
  \centering
  \includegraphics[width=0.879\textwidth]{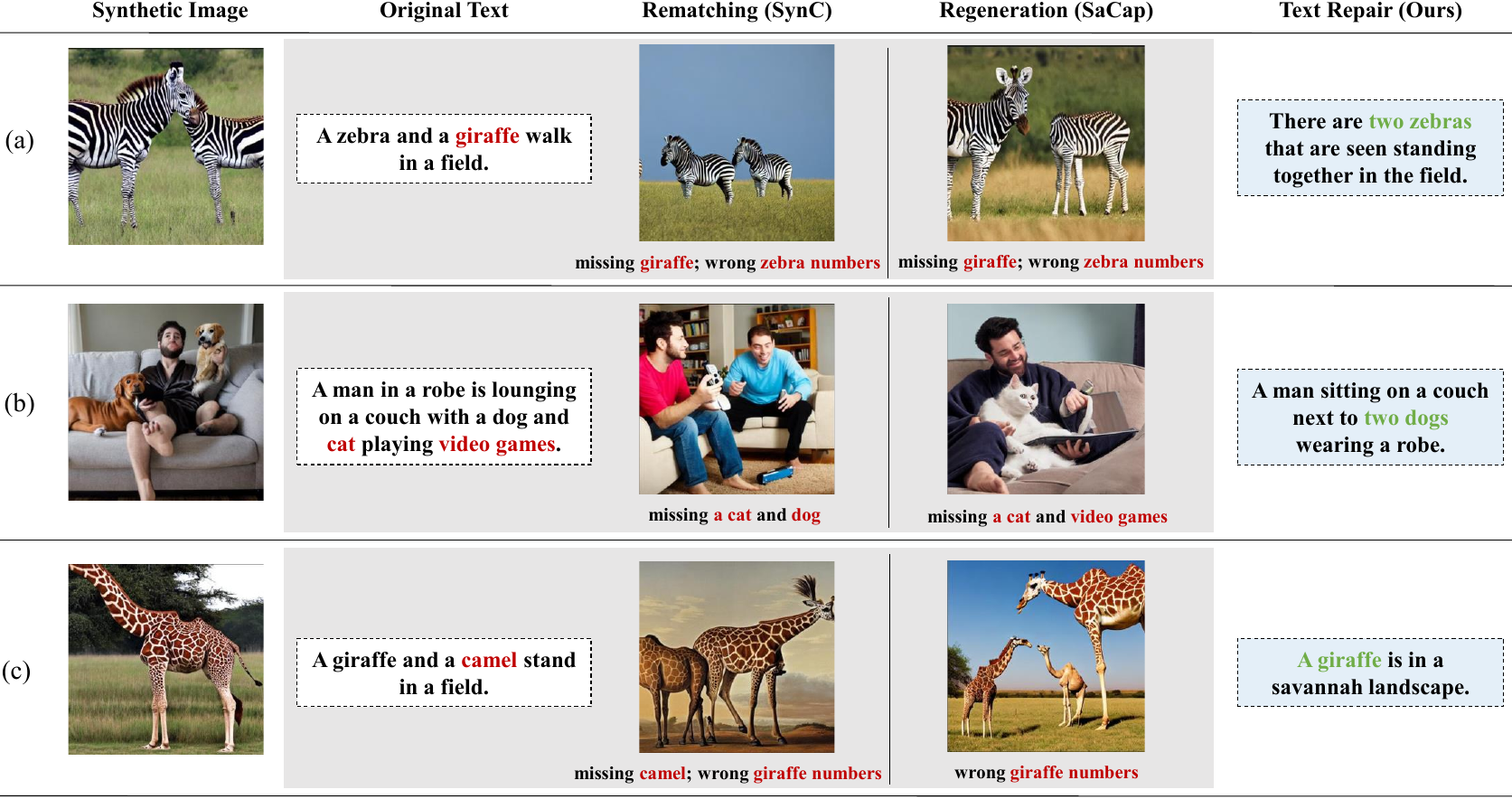}
  \caption{Visual comparison of realignment strategies. Red text marks unsupported concepts. Global-similarity rematching and regeneration miss concepts or introduce inconsistencies, while our method aligns text with the image.}
  \label{fig:wide_figure_7}
\end{figure*}

\noindent\textbf{Text Embedding Distributions.} To examine whether entity-guided rewriting changes the global semantic characteristics of the training corpus, we visualize text embeddings using Principal Component Analysis~\cite{shlens2014pca}. Fig.~\ref{fig:wide_figure_5}(a) shows the distributions of two stylistically distinct datasets, MSCOCO and Flickr30k, in the text embedding space. The orange and blue points denote MSCOCO and Flickr30k captions, respectively, and their clear separation reflects the different linguistic styles of the two datasets. Fig.~\ref{fig:wide_figure_5}(b) further compares text embeddings before and after rewriting, where orange points represent the original texts and green points represent the rewritten ones. The overall distribution remains largely unchanged after rewriting, indicating that ReCap does not distort the global semantic structure of the corpus. Instead, it mainly acts as a pair-level supervision repair mechanism, improving the alignment between synthetic images and captions while preserving the original distributional characteristics of the data.

\noindent\textbf{Visual Comparison with Realignment Methods.}
We conduct a visual analysis of different realignment strategies, as shown in Fig.~\ref{fig:wide_figure_7}. In case (a), the synthetic image lacks the giraffe concept, which neither rematching nor regeneration can recover. Rematching yields semantically incomplete pairs, while regeneration fails to introduce the missing entity. In case (b), regeneration introduces the missing cat but removes the dog, revealing a trade-off between adding missing content and preserving valid content, whereas rematching cannot recover the missing semantics. In case (c), neither method can generate both giraffe and camel simultaneously, highlighting the difficulty of recovering multiple missing concepts. Unlike global rematching or regeneration, ReCap preserves the synthetic image and directly repairs fine-grained caption errors to produce more faithful image--text pairs.

\noindent\textbf{Error Analysis of Entity Presence Detector.} The entity presence detector verifies whether entities extracted from the conditional text are present in the image, with predictions restricted to the extracted entity set. This constrained formulation avoids open-ended entity generation and explicit normalization of lexical variants or coreferential expressions. False positives merely retain unsupported entities from the conditional text and therefore introduce no additional entities.
To assess robustness to false-negative noise and the effect of ADWL under such conditions, we simulate noise by randomly removing correctly verified entities. As shown in Table~\ref{tab:entity_ablation}, our framework remains resilient to error propagation: even under extreme noise settings, the selection and filtering strategy preserves the original text supervision and maintains performance comparable to the baseline. ADWL further alleviates the impact of noisy samples, validating its effectiveness under unreliable supervision.

\noindent\textbf{Entity Recall Evaluation on the Test Set.} We evaluate whether entity-faithful supervision repair improves entity preservation in generated captions. Following this goal, we use NLTK to extract entities from captions generated by the baseline, CapDec, DeCap, ReCap, and ReCap trained on real image-text pairs, and compute their recall with respect to the ground-truth entity set. As shown in Table~\ref{tab:entity_recall}, on MSCOCO, ReCap improves entity recall from 0.3399 to 0.3510. This supports the effectiveness of our entity-guided repair strategy in improving the preservation of visually grounded entities.

\noindent\textbf{Data Preprocessing and Preparation Cost.} We compare the data-preparation time of ReCap with SynC and SaCap in Table~\ref{tab:time_cost}, with all experiments conducted on a single RTX 3090 GPU. On MSCOCO, ReCap requires 159.5 hours in total, which is slightly higher than SynC (141.3 hours) but significantly lower than SaCap (549 hours), achieving approximately a 3.4$\times$ speedup over SaCap. In terms of per-image time cost, ReCap takes 1.01 seconds, comparable to SynC (0.90 seconds) and substantially faster than SaCap (3.49 seconds). Compared to existing methods that rely on costly image regeneration~\cite{liu2025stacap} or are constrained by data distribution~\cite{kim2025sync}, our approach optimizes alignment via low-cost text rewriting, improving efficiency, scalability, and semantic consistency.

\section{Conclusion}
We present ReCap, a plug-and-play framework for zero-shot image captioning that repairs structured fine-grained misalignment in synthetic supervision. By combining entity-guided text rewriting with adaptive dynamic-weighted learning, ReCap improves supervision fidelity and reduces residual noise in synthetic image-text pairs, leading to more accurate and transferable caption generation. Extensive experiments show that repairing synthetic supervision at the pair level, rather than relying solely on rematching or regeneration based on the global similarity, substantially improves both in-domain and cross-domain zero-shot captioning performance. These findings highlight the importance of structured supervision repair for robust zero-shot captioning.

\begin{acks}
This work was supported by the National Natural Science Foundation of China (No. 62406081), the Guangxi Natural Science Foundation (No. 2025GXNSFBA069232), and the Guangxi Bagui Youth Talent Program.
\end{acks}

\bibliographystyle{ACM-Reference-Format}
\bibliography{main}

\clearpage
\appendix

\section{Prompt}
To ensure entity-faithful supervision repair, we explicitly constrain the large language model to rewrite captions according to the image-supported entity set $E'$ and the retrieval caption for repair, while suppressing unsupported or misleading content. The prompt used for caption rewriting is shown below:
\begin{tcolorbox}[title=, colback=gray!5, colframe=black!30]
\footnotesize
Caption: <caption>

Detected Entities: <E'>

- Rewrite the caption using only the entities listed in ``Detected entities''. 

- All detected entities must appear in the rewritten caption. 

- You may add minimal, reasonable details such as entity positions, simple actions, or basic scene context to make the description coherent, but all content must be grounded in the provided caption and the detected entity information. Include entity names, attributes (e.g., color, position, action), and relationships only if they are present in the caption or explicitly detected. **Do not introduce any information beyond what is provided**.

- Do not assume any interactions between entities that are not naturally implied.  

- Keep the sentence fluent, coherent, and grammatically correct. 

- Preserve the style and tone of the caption as much as possible, but avoid copying incorrect or misleading information. 

- Output only one complete sentence without explanations or labels.
\end{tcolorbox}

\section{More Ablation Studies}
Table~\ref{tab:results} evaluates the contributions of the major components in ReCap and compares our structured repair pipeline with direct rewriting based on vision-language models (VLMs). Rewriting using only the original text (OTR) leads to a performance degradation. This suggests that, without introducing additional grounding information, attribute errors and entity omissions in the original captions may be preserved or even amplified, resulting in overly simplified descriptions that deviate from the semantic distribution of the original corpus. Similarly, removing the DINO-based entity detection and verification module (w/o GD) reduces the CIDEr score from 98.5 to 79.8, demonstrating that explicit entity annotations are crucial for identifying and correcting subtle mismatches in synthetic image--text pairs. The variants without candidate construction (w/o CC) or candidate selection (w/o Select) outperform the original baseline but remain consistently inferior to the complete ReCap framework. These results indicate that both targeted candidate construction and generate-then-align selection are necessary for producing reliable rewritten captions, rather than directly accepting potentially noisy outputs from a language model.

We further compare ReCap with direct VLM-based rewriting approaches. The VLM variant, using InstructBLIP-Vicuna-7B\footnote{InstructBLIP: Towards General-purpose Vision-Language Models with Instruction Tuning, NeurIPS 2023.}, generates a new caption directly from the synthetic image, whereas VLM+GD additionally uses the detected and verified entities as guidance. Although entity guidance improves the B@4 and CIDEr scores over direct VLM generation, both variants remain substantially inferior to ReCap, particularly in terms of CIDEr. This suggests that direct VLM rewriting may produce descriptions that are visually plausible but deviate from the semantics, phrasing patterns, or data distribution of the original corpus. In contrast, ReCap preserves reliable content from the original caption, identifies missing or inconsistent entities, constructs targeted rewriting candidates, and selects the candidate that best retains the image-grounded semantics. Consequently, the complete framework achieves the best performance across all four evaluation metrics. These results confirm that the effectiveness of ReCap does not arise from merely combining entity detection with text generation. Instead, it stems from the coordinated integration of entity detection, constrained rewriting, and semantic selection, which explicitly repairs structured, fine-grained misalignments in synthetic supervision.

As shown in Table~\ref{tab:results}, directly replacing the original texts with captions generated by a vision-language model leads to clear performance degradation. These generated captions may deviate from the semantic and stylistic characteristics of the original corpus. Moreover, naively stacking entity detection and rewriting without the full repair and weighting design does not improve performance. In contrast, the full ReCap framework achieves clear gains by explicitly correcting entity omission and fine-grained semantic inconsistencies in synthetic supervision. These results show that the effectiveness of ReCap does not come from simply combining off-the-shelf components, but from explicitly modeling and repairing structured fine-grained misalignment.

\begin{table}[t]
\centering
\caption{Detailed ablation analysis of entity detection and text rewriting.}
\label{tab:results}
\begin{tabular}{lcccc}
\toprule
\textbf{Method} &\textbf{B@4} & \textbf{M} & \textbf{C} & \textbf{S} \\
\midrule
Baseline & 26.1 & 24.4 & 89.8 & 18.0 \\
\text{+OTR} & 21.7 & 22.5 & 87.6 & 16.5 \\ 
\text{w/o GD} & 21.9 & 22.9 & 79.8 & 17.2 \\ 
\text{w/o CC} & 28.2 & 24.6 & 94.5 & 18.2 \\
\text{w/o Select} & 28.0 & 24.7 & 94.6 & 18.4 \\ 
\text{VLM} & 20.7 & 25.1 & 71.8 & 18.3 \\ 
\text{VLM+GD} & 22.8 & 23.6 & 80.7 & 18.3 \\
ReCap & \textbf{29.3} & \textbf{25.2} & \textbf{98.5} & \textbf{18.8} \\
\bottomrule
\end{tabular}
\end{table}

\begin{table}[t]
\centering
\caption{Entity verification performance on real and synthetic images.}
\label{tab:recall}
\begin{tabular}{lccc}
\toprule
Image type & Recall & Precision & F1 \\
\midrule 
Real images & 0.78 & 0.80 & 0.79 \\
Synthetic images   & \textbf{0.89} & \textbf{0.93} & \textbf{0.91} \\
\bottomrule
\end{tabular}
\end{table}

\section{Analysis of the Entity Presence Detector}
\noindent\textbf{Fairness.} We emphasize that the entity presence detector used in ReCap is not equivalent to a conventional object detector. It operates only on a predefined set of queried entities extracted from the source text, and therefore cannot be used for open-set object discovery in image captioning. Moreover, it is used only during the synthetic supervision repair stage and is not involved in caption model training or inference.

\noindent\textbf{Applicability to Synthetic Images.} We evaluate the suitability of the entity presence detector for synthetic images by randomly sampling 100 synthetic images and measuring its recall, precision, and F1 score, as reported in Table~\ref{tab:recall}. The detector achieves consistently better performance on synthetic images than on real images across all three metrics. In particular, the higher recall indicates that a larger proportion of the queried entities can be successfully localized, while the higher precision suggests that the detected regions are also less likely to contain irrelevant objects. Consequently, the improved F1 score demonstrates a better overall balance between detection coverage and reliability. One possible explanation is that synthetic images are generated from textual descriptions, causing the queried entities to appear more explicitly and occupy relatively salient regions. In contrast, real-world images often contain more complex scenes, smaller objects, background clutter, and heavier occlusion, which make entity localization more challenging. These results suggest that Grounding DINO provides sufficiently accurate and reliable entity annotations for our synthetic supervision repair setting.

\section{Analysis of Failure Cases}
\begin{figure*}[!t]
  \centering
  \includegraphics[width=1.0\textwidth]{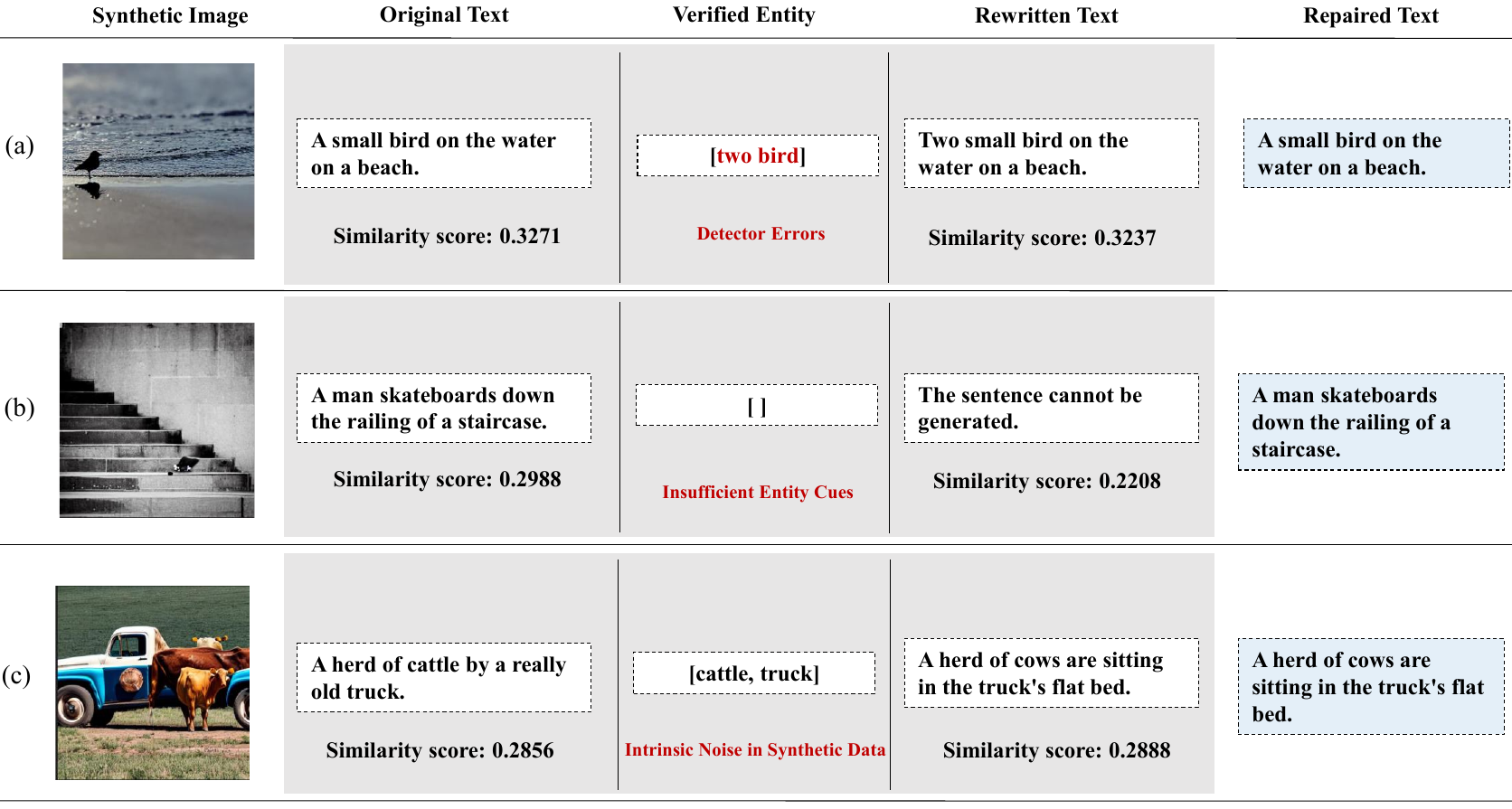}
  \caption{Failure Cases.}
  \label{fig:wide_figure_15}
\end{figure*}

\noindent\textbf{Detection Errors.} In Fig.~\ref{fig:wide_figure_15}(a), the detector incorrectly verifies the presence of two birds, which may mislead the rewriting module into introducing unsupported entities. Such false-positive entity annotations can propagate through the pipeline and reduce the semantic faithfulness of the repaired caption. In this case, our candidate selection step, which evaluates both the rewritten candidates and the original text using the global similarity, helps suppress erroneous rewrites and preserves valid information from the source caption. By retaining candidates that remain semantically consistent with the original caption, the selection module provides an additional safeguard against unreliable entity guidance. This shows the sensitivity of the repair pipeline to upstream errors and the role of our generate-then-align strategy in reducing their impact.

\noindent\textbf{Insufficient Entity Cues.} In Fig.~\ref{fig:wide_figure_15}(b), when the detector fails to provide useful entity information, the text rewriting process lacks sufficient cues for the language model to generate a meaningful caption. As a result, the language model may generate empty or uninformative outputs. ReCap mitigates the impact of such failures by retaining the original caption as a candidate during the generate-then-align stage, thereby reducing the risk of harmful rewrites and preventing unstable language model outputs from corrupting the final supervision. At the same time, our ADWL mechanism effectively suppresses the model from learning incorrect mappings by reducing the learning weight of erroneous samples.

\noindent\textbf{Intrinsic Noise in Synthetic Data.} 
In Fig.~\ref{fig:wide_figure_15}(c), a mixture of multiple entities (i.e., cows and trucks) exhibits irreducible noise inherent to synthetic image generation. Such noise cannot always be fully resolved by text repair alone. In these cases, our adaptive dynamic weighted learning mechanism reduces the contribution of unreliable pairs during training, allowing the model to focus on more faithful supervision. That is, while entity-guided rewriting repairs many structured fine-grained errors, residual noisy pairs are further handled through reliability-aware weighting.

\section{Training and Inference Efficiency}

We analyze the efficiency of ReCap in training and inference. Since we improve zero-shot captioning mainly through synthetic supervision repair rather than inference-time retrieval or external memory, it is necessary to evaluate the efficiency of the captioning pipeline.
\begin{table}[t]
\centering
\caption{Comparison of training overhead.}
\begin{tabular}{l|cc}
\toprule
\textbf{Method}  & \textbf{MSCOCO Training} & \textbf{Flickr30k Training}  \\
\midrule
CapDec    & 4.9h & 1.4h \\
VIECap   & 5.1h & 1.5h \\
SynTIC  & 4.7h & 1.2h \\
ReCap  & 1.3h & 0.7h \\
\bottomrule
\end{tabular}
\label{tab:train_time}
\end{table}

\begin{table}[t]
\centering
\caption{Comparison of inference overhead.}
\begin{tabular}{l|ccc}
\toprule
\textbf{Method}  & \textbf{Retrieval/Detection} & \textbf{Decoding} & \textbf{Total Time} \\
\midrule 
CapDec   & -- & 186.7ms & 186.7ms \\
VIECap  & 0.62ms & 127.9ms & 128.6ms \\
SynTIC  & 61.3ms & 45.3ms  &  106.6ms\\
ReCap  & -- & 85.3ms  & 85.3ms \\
\bottomrule
\end{tabular}
\label{tab:inference_time}
\end{table}
\noindent\textbf{Training Efficiency.}
Table~\ref{tab:train_time} compares the training time of ReCap with representative zero-shot captioning baselines on MSCOCO and Flickr30k, including CapDec, VIECap, and SynTIC. The results show that ReCap remains computationally efficient during training, while achieving stronger performance. This suggests that repairing synthetic supervision provides a practical way to improve learning effectiveness without introducing excessive optimization overhead.

\noindent\textbf{Inference Efficiency.}
Table~\ref{tab:inference_time} reports the per-image inference cost of different methods. ReCap achieves the lowest overall latency, requiring only 85.3ms. Compared with CapDec, ReCap reduces decoding time by more than 50\%, indicating that its captioning pipeline is computationally lighter at inference time. Although VIECap uses only a lightweight retrieval module (0.62ms), its total latency remains higher because of a slower decoding stage. SynTIC adopts a lightweight decoder, but its inference still incurs extra overhead because it requires object detection for each input image. In contrast, ReCap does not rely on retrieval, detection, external memory, or entity verification during inference, resulting in a simpler and more streamlined generation pipeline.

Overall, these results show that ReCap achieves strong practical efficiency. While supervision repair is performed offline during data preparation, the resulting captioning model introduces no extra inference-time retrieval or detection overhead, and improves inference speed compared with existing zero-shot captioning baselines.

\section{More Experiments and Visualization}
\noindent\textbf{Comparison of Different Weighting Strategies.} We compare our adaptive dynamic weighted learning strategy (ADWL) with two simpler weighting schemes: entity-coverage weighting and CLIP-based weighting. Entity-coverage weighting estimates the reliability of a repaired image-caption pair according to the proportion of image-grounded entities preserved in the rewritten caption. Although this strategy can penalize captions that omit salient entities, it relies primarily on explicit entity matching and therefore cannot adequately capture incorrect attributes, erroneous relations, or semantically inconsistent descriptions. Moreover, captions with similar entity-coverage scores may differ substantially in their overall quality, since merely mentioning the correct entities does not guarantee that they are described accurately or coherently. CLIP-based weighting instead uses global image-text similarity to provide a coarse estimate of semantic alignment. However, global similarity is often insensitive to fine-grained errors in synthetic supervision, such as omitted entities, incorrect attributes, erroneous relations, or subtle wording mismatches. Consequently, CLIP-based weighting may assign similar importance to pairs that are globally plausible but locally misleading. In contrast, ADWL estimates the reliability of each repaired pair by comparing the rewritten caption with multiple retrieved textual references in the text semantic space. This design captures richer contextual and semantic consistency beyond explicit entity coverage or global image-text alignment, providing a more stable and fine-grained estimate of whether the repaired caption preserves the main image-grounded semantics. As shown in Table~\ref{tab:clip_vs}, entity-coverage weighting yields only limited improvements over the baseline, while CLIP-based weighting performs better by incorporating global semantic information. ADWL achieves the strongest overall performance, demonstrating its greater effectiveness in downweighting residual noisy pairs during training.

\begin{table}[t]
\caption{Comparison of different weighting strategies on the repaired training data.}
\centering
\begin{tabular}{l|cccc}
\toprule
\textbf{Method}   &\textbf{B@4} & \textbf{M} & \textbf{C} & \textbf{S} \\
\midrule 
Baseline (Repaired Data)         &27.7&\textbf{25.2}&95.6&18.6 \\
{\textit{+Entity-coverage  Weighting}} & 28.0 & 25.0 & 96.1 & 18.4\\
{\textit{+CLIP-based Weighting}}       &29.1&25.1&97.7&18.7\\
{\textit{+ADWL}}      & \textbf{29.3} & \textbf{25.2} & \textbf{98.5}  & \textbf{18.8} \\
\bottomrule
\end{tabular}
\label{tab:clip_vs}
\end{table}

\noindent\textbf{Controlled Comparison with Other Realignment Methods.}
To further isolate the effect of supervision repair itself, we compare different realignment strategies under the same baseline captioning model. Specifically, we remove ADWL and then train the same baseline separately on synthetic pairs refined by SynC, SaCap, and our rewriting-based repair strategy. Table~\ref{tab:repairs_domain} reports the in-domain results. Under the same captioning backbone, ReCap achieves the best overall performance, obtaining the highest scores on B@4, CIDEr, and METEOR. Compared with the unrepaired baseline, ReCap yields consistent gains across all metrics, including a notable improvement of +3.2 CIDEr. This suggests that rewriting-based supervision repair produces more faithful training pairs than image rematching or regeneration. Table~\ref{tab:repairs_cosdomain} reports the cross-domain results. ReCap consistently outperforms other realignment strategies across all splits. Compared with the strongest competing method, SaCap, it improves performance by +0.7, +1.2, +2.9, and +1.7, with the largest gain on the Out split. These results indicate that ReCap provides more transferable supervision and stronger generalization to novel visual concepts.

\begin{table}[t]
\caption{Comparison of different realignment strategies on MSCOCO under the same baseline model.}
\centering
\begin{tabular}{l|cccc}
\toprule
\textbf{Method}  &\textbf{B@4} & \textbf{M} & \textbf{C} & \textbf{S} \\
\midrule 
Baseline         &26.1&24.4&89.8&18.0\\
\textit{+SynC}    &26.3&24.5&91.0&18.1\\
\textit{+SaCap}     &25.8&24.5&91.5&\textbf{18.4}\\
\textit{+ReCap} &\textbf{26.6}&\textbf{24.6}&\textbf{93.0}&\textbf{18.4} \\
\bottomrule
\end{tabular}
\label{tab:repairs_domain}
\end{table}

\begin{table}[t]
\caption{Comparison of different realignment strategies on MSCOCO $\rightarrow$ NoCaps under the same baseline model. }
\centering
\begin{tabular}{l|cccc}
\toprule
\textbf{Method}  & \textbf{In} & \textbf{Near} & \textbf{Out} & \textbf{All} \\
\midrule 
Baseline         &62.9&60.3&42.4&56.5\\
\textit{+SynC}         &64.2&60.8&45.2&57.7\\
\textit{+SaCap}       &65.3&61.5&45.6&58.4\\
\textit{+ReCap} &\textbf{66.0}&\textbf{62.7}&\textbf{48.5}&\textbf{60.1}\\
\bottomrule
\end{tabular}
\label{tab:repairs_cosdomain}
\end{table}

\begin{table}[t]
\caption{Validation on SS1M $\rightarrow$ MSCOCO.}
\centering
\begin{tabular}{l|cccc}
\toprule
\textbf{Method}  &\textbf{B@4} & \textbf{M} & \textbf{C} & \textbf{S} \\
\midrule 
SynC  & 11.3 & 14.2 & 49.1 & 11.3 \\
ReCap+retrieval & \textbf{12.7} & \textbf{17.4} & \textbf{52.3} & \textbf{13.6} \\
\bottomrule
\end{tabular}
\label{tab:ss1m}
\end{table}

\begin{table}[t]
\centering
\caption{Human evaluation.}
\label{tab:comparison11}
\begin{tabular}{l | c c c}
\toprule
\textbf{Metric} & Baseline & ReCap & $\Delta$ \\
\midrule
Entity Accuracy     & 3.76 & 4.00 & $+0.24$ \\
Attribute Binding   & 3.88 & 3.95 & $+0.07$ \\
Hallucination-free  & 3.61 & 3.94 & $+0.33$ \\
Overall Quality     & 3.75 & 3.96 & $+0.21$ \\
\bottomrule
\end{tabular}
\end{table}

\noindent\textbf{Validation on Web-Sourced Data.}
As shown in Table~\ref{tab:ss1m}, ReCap+retrieval consistently outperforms SynC across all metrics when trained on the web-sourced SS1M\footnote{Unsupervised Image Captioning, CVPR 2019.} dataset and evaluated on MSCOCO. This demonstrates that repairing entity omissions and semantic inconsistencies improves the quality of web-crawled synthetic supervision. 

\noindent\textbf{Human Evaluation Results.}
We conduct a human evaluation on the MSCOCO test set. Specifically, we invite three graduate students to perform an evaluation of 100 randomly sampled cases. Captions generated by ReCap and the baseline using original synthetic data are scored on a scale of 1–5 based on Entity Accuracy (whether all mentioned objects are present), Attribute Binding (whether attributes are correctly associated), and Hallucination-Free (whether the caption avoids introducing non-existent entities). As shown in Table~\ref{tab:comparison11}, ReCap outperforms the baseline in fine-grained accuracy.

\begin{figure*}[t]
  \centering
  \includegraphics[width=\textwidth]{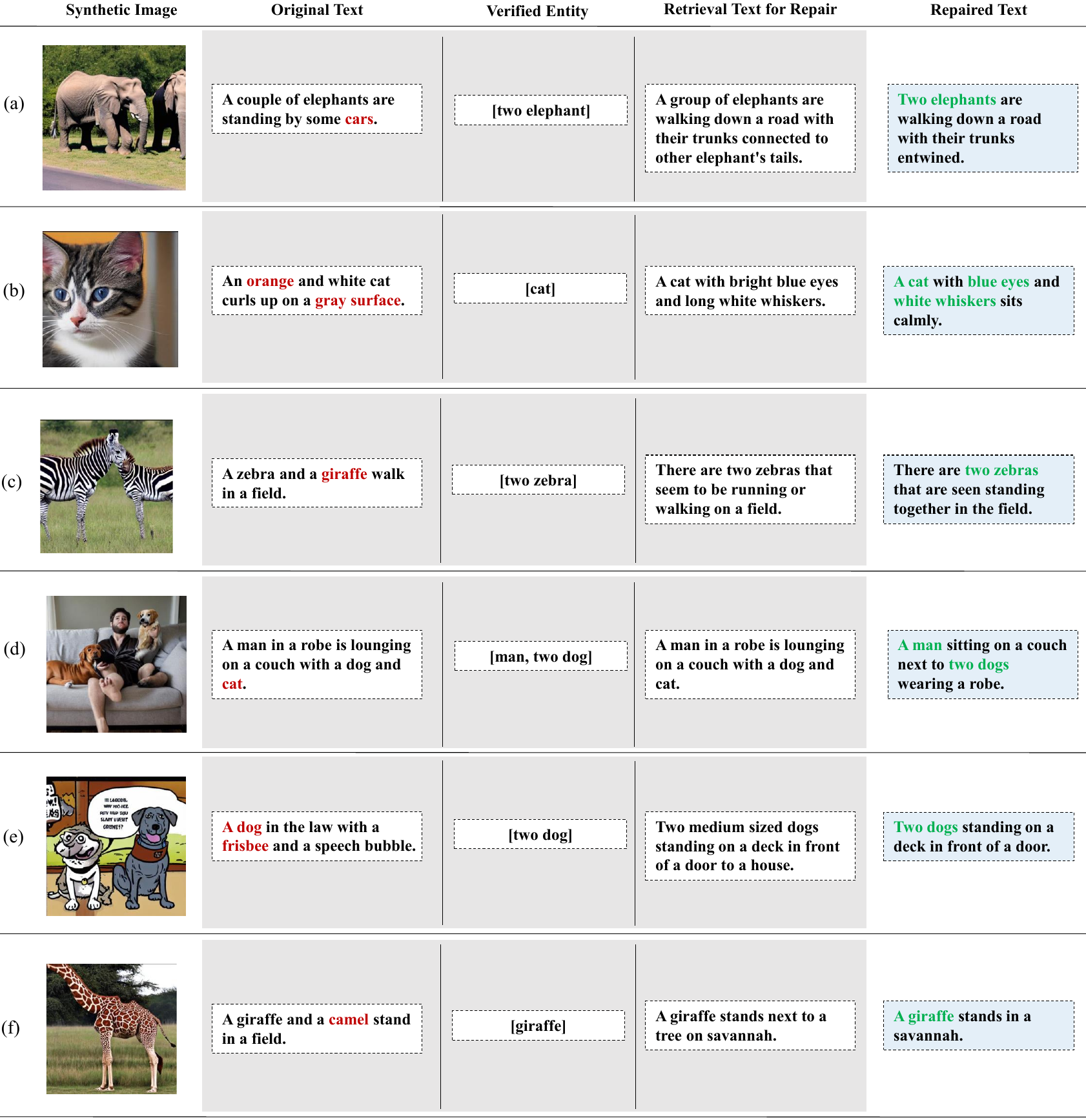}
  \caption{Qualitative examples of supervision repair in ReCap. In each case, the synthetic image and the original text are taken from the same initial synthetic image-text pair. The retrieval text for repair is the globally most similar text to the synthetic image and serves as the basis for language-model rewriting. Red text marks content in the original text that is inconsistent with the synthetic image, while green text marks corrected content in our repaired caption that is faithful to the image.}
  \label{fig:wide_figure_4}
  
\end{figure*}

\noindent\textbf{Qualitative Evaluation of Repair.}
We present representative examples of synthetic supervision before and after repair in Fig.~\ref{fig:wide_figure_4}. The synthetic images fail to faithfully realize parts of the original text and exhibit several forms of structured fine-grained misalignment, including missing entities (e.g., missing car, giraffe, cat, frisbee, and camel in (a), (c), (d), (e), and (f), respectively), incorrect attribute grounding (e.g., wrong cat attributes in (b)), and incorrect quantity (e.g., wrong number of zebras in (c) and dogs in (e)). In contrast, the repaired captions are more faithful to the corresponding synthetic images and better preserve image-supported entities, attributes, and scene details. Our repair process does not directly rewrite the original caption. Instead, ReCap first retrieves the text that is globally most similar to the synthetic image, and then refines this retrieval text under entity constraints to obtain a repaired caption that is more faithful to the image. These examples qualitatively illustrate that ReCap repairs supervision errors at the caption level, rather than merely improving global plausibility.

\begin{figure*}[t]
  \centering
  \includegraphics[width=1.0\textwidth]{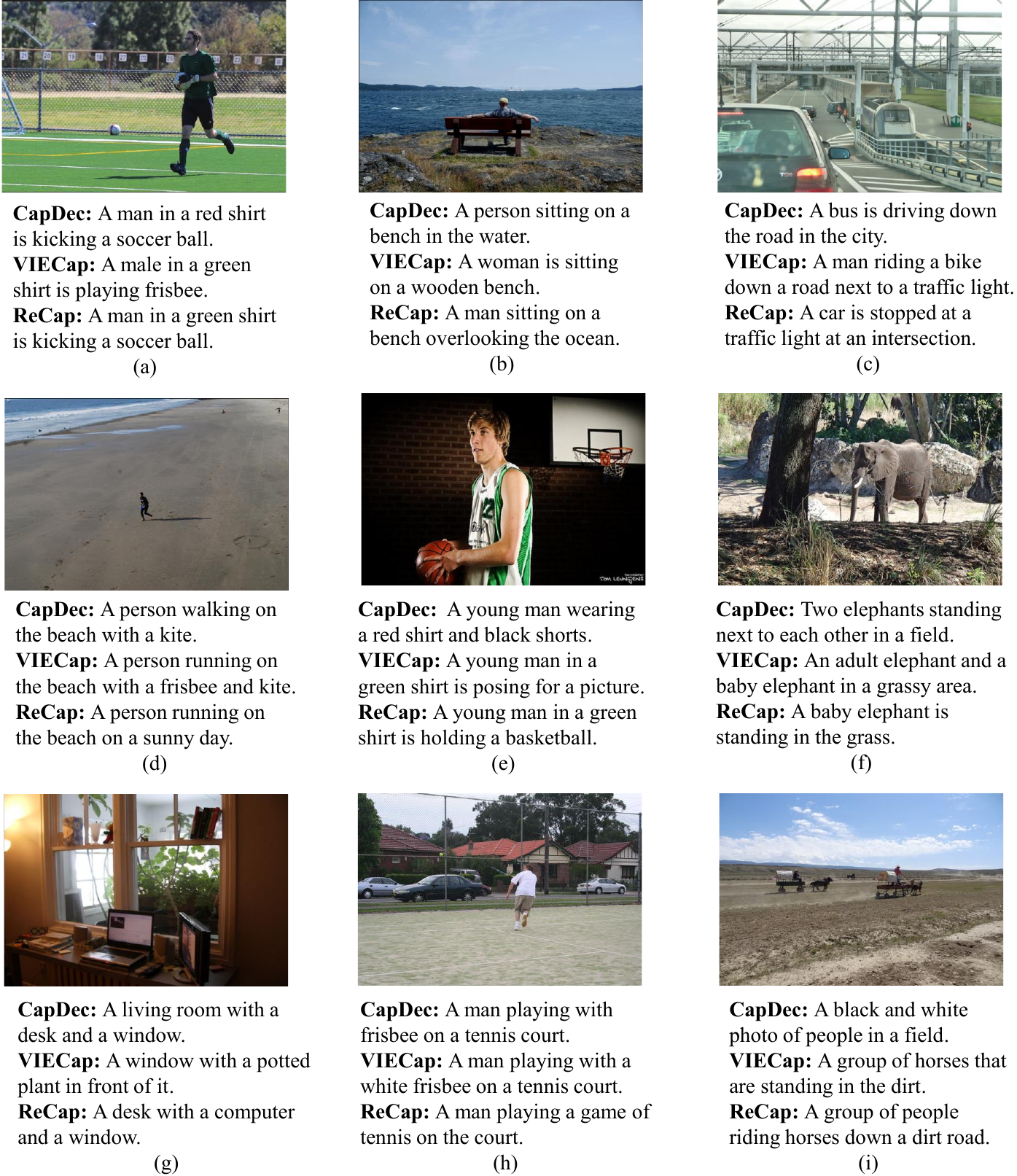}
  \caption{Captioning results of different zero-shot image captioning methods.}
  \label{fig:wide_figure_6}
\end{figure*}

\noindent\textbf{Captioning Results.} We compare the captions generated by ReCap with those produced by the open-source baselines, CapDec and VIECap, on MSCOCO, as shown in Fig.~\ref{fig:wide_figure_6}. Across these examples, the baselines frequently exhibit fine-grained errors, including incorrect attributes, mistaken object identities, unsupported objects, and incomplete understanding of object relations or scene details.

In particular, CapDec often produces coarse or inaccurate descriptions, such as predicting the wrong clothing color in (a), describing an unnatural relation in (b), misidentifying the main object in (c), or missing important scene details in (e) and (g). VIECap, while sometimes capturing part of the scene semantics, is more prone to semantic drift or unsupported content, such as mistaking the activity in (a) and (h), misclassifying gender in (b), generating a completely irrelevant scene in (c), or introducing unsupported objects and relations in (d), (f), and (i). In contrast, ReCap consistently produces captions that are more faithful to the visual content, better capturing fine-grained attributes, object identities, and relations. It also avoids several severe errors made by the baselines, such as hallucinated objects in (d) and (f), missing key objects in (e), or incorrect interactions in (i). Overall, these examples show that ReCap yields more precise, detailed, and semantically consistent captions than CapDec and VIECap. This qualitative improvement is consistent with our central claim: by improving the fidelity of synthetic supervision during training, ReCap enables the captioning model to better capture subtle visual details and fine-grained relationships at inference time.

\end{document}